# Multi-Task Feature Learning Via Efficient $\ell_{2,1}$-Norm Minimization

**Jun Liu**
Department of CSE
Arizona State University
Tempe, AZ 85287, USA
J.Liu@asu.edu

**Shuiwang Ji**
Department of CSE
Arizona State University
Tempe, AZ 85287, USA
Shuiwang.Ji@asu.edu

**Jieping Ye**
Department of CSE
Arizona State University
Tempe, AZ 85287, USA
Jieping.Ye@asu.edu

## Abstract

The problem of joint feature selection across a group of related tasks has applications in many areas including biomedical informatics and computer vision. We consider the $\ell_{2,1}$-norm regularized regression model for joint feature selection from multiple tasks, which can be derived in the probabilistic framework by assuming a suitable prior from the exponential family. One appealing feature of the $\ell_{2,1}$-norm regularization is that it encourages multiple predictors to share similar sparsity patterns. However, the resulting optimization problem is challenging to solve due to the non-smoothness of the $\ell_{2,1}$-norm regularization. In this paper, we propose to accelerate the computation by reformulating it as two equivalent smooth convex optimization problems which are then solved via the Nesterov's method—an optimal first-order black-box method for smooth convex optimization. A key building block in solving the reformulations is the Euclidean projection. We show that the Euclidean projection for the first reformulation can be analytically computed, while the Euclidean projection for the second one can be computed in linear time. Empirical evaluations on several data sets verify the efficiency of the proposed algorithms.

## 1 Introduction

Multi-task learning [1, 2, 3, 7, 17, 18, 19, 26, 28] has recently received increasing attention in machine learning, artificial intelligence, and computer vision. It aims to learn the shared information among related tasks for improved performance, and has been successfully employed in applications including medical diagnosis [3, 18, 26], handwritten character recognition [17, 18], conjoint analysis [1, 28], and text classification [28].

The underlying assumption behind many multi-task learning algorithms is that the tasks are related to each other. Thus, a key issue is how to capture the task relatedness and take this into account in the learning formulation [2]. Existing multi-task learning formulations employ various strategies to capture the task relatedness.

Evgeniou and Pontil [7] assumed that all the tasks are related so that the true models (for all the tasks) are close to a given (unknown) common model. They proposed to solve the multi-task models by the $\ell_2$-norm regularization analogous to SVMs [4]. Obozinski et al. [17, 18, 19] proposed to penalize the sum of the $\ell_2$-norms of the blocks of coefficients associated with each feature across tasks, leading to an $\ell_{2,1}$-norm regularized nonsmooth convex optimization problem. Obozinski et al. proposed to solve the problem by the blockwise boosting scheme [29] that follows the regularization path. However, there is no known convergence rate for the blockwise boosting scheme. Argyriou et al. [1] assumed that the tasks share a small subset of features (via the feature matrix), and formulated the problem as the squared $\ell_{2,1}$-norm regularized non-convex optimization problem. If the feature matrix is not learned (set to an identity matrix), the problem reduces to the squared $\ell_{2,1}$-norm regularized convex optimization problem, which is similar to the one proposed in [17, 18]. Xiong et al. [26] proposed a probabilistic framework by imposing the automatic relevance determination prior on the parameters across different tasks. Zhang et al. [28] proposed a unified probabilistic framework, where the task parameters share a common structure through latent variables.

Multi-task feature learning via the $\ell_{2,1}$-norm regularization has been studied in [1, 17, 18, 19]. One appealing property of the $\ell_{2,1}$-norm regularization is that it encourages multiple predictors from different tasks to share similar parameter sparsity patterns. The $\ell_{2,1}$-norm regularization has also been successfully employed in the group Lasso [27], the logistic group-lasso [13], and the generalized linear models [21]. When the loss is convex, the $\ell_{2,1}$-norm regularized problem is convex, and admits a globally optimal solution. Previous approaches [1, 17, 18, 19] for solving the $\ell_{2,1}$-norm regularized problem apply an iterative procedure, which may converge slowly. To the best of our knowledge, no known global convergence rate has been established for these algorithms.

In this paper, we propose to solve the $\ell_{2,1}$-norm regularized problem using the first-order black-box methods [14, 16], which evaluate at each iteration the function value and (sub)gradient only. One major challenge lies in the nonsmoothness of the $\ell_{2,1}$-norm regularization term. It has been shown [14, 15] that the lower complexity bound for smooth convex optimization is significantly better than that of nonsmooth convex optimization. Moreover, Nesterov's method [14, 16] is an optimal first-order black-box method for smooth convex optimization.

Due to the superior convergence rate of the smooth convex optimization over the nonsmooth one, we propose to reformulate the non-smooth $\ell_{2,1}$-norm regularized problem as its equivalent smooth convex optimization problem. More specifically, we consider two smooth reformulations in this paper: 1) the reformulation by the introduction of additional variables; and 2) the $\ell_{2,1}$-ball constrained optimization. When applying Nesterov's method for solving both reformulations, a key building block is the Euclidean projection onto the constraint. We show that the Euclidean projection for the first reformulation can be computed analytically, while the projection for the second one can be computed in linear time. With the equivalent reformulations, we can solve the $\ell_{2,1}$-norm regularization problem with a smooth loss function in a time complexity of $O(\frac{1}{\sqrt{\varepsilon}}(mn+nk))$, where $m$, $n$, $k$ and $\varepsilon$ denote the total number of training samples, the sample dimensionality, the number of tasks, and the desired accuracy, respectively. Empirical evaluations on several data sets verify the efficiency of the proposed algorithms.

**Notations:** Scalars are denoted by lower case italic letters, vectors by lower case bold face letters, and matrices by capital italic letters. Let $\|\cdot\|$ denote the Euclidean ($\ell_2$) norm, $\|\cdot\|_1$ the $\ell_1$-norm, $\|\cdot\|_F$ the Frobenius norm of a matrix, and $\|\cdot\|_{2,1}$ the $\ell_{2,1}$-norm of a matrix.

**Organization:** In Section 2, we introduce the $\ell_{2,1}$-norm regularized multi-task feature learning in the probabilistic framework; in Section 3, we present the proposed method for solving the equivalent smooth reformulations; in Section 4, we detail how to efficiently solve the Euclidean projection problems; in Section 5, we report empirical results; and we conclude this paper in Section 6.

## 2 A Probabilistic Framework for Multi-task Feature Learning

In multi-task learning, we are given a training set of $k$ tasks $\{(\mathbf{a}_i^j, y_i^j)\}_{i=1}^{m_j}$, $j=1,2,\ldots,k$, where $\mathbf{a}_i^j \in \mathbb{R}^n$ denotes the $i$-th training sample for the $j$-th task, $y_i^j$ denotes the corresponding output, $m_j$ is the number of training samples for the $j$-th task, and $m = \sum_{j=1}^k m_j$ is the total number of training samples (including possible duplicates). Let $A_j = [\mathbf{a}_1^j, \ldots, \mathbf{a}_{m_j}^j]^T \in \mathbb{R}^{m_j \times n}$ denote the data matrix for the $j$-th task, $A = [A_1^T, \ldots, A_k^T]^T \in \mathbb{R}^{m \times n}$, $\mathbf{y}_j = [y_1^j, \ldots, y_{m_j}^j]^T \in \mathbb{R}^{m_j}$, and $\mathbf{y} = [\mathbf{y}_1^T, \ldots, \mathbf{y}_k^T]^T \in \mathbb{R}^m$. In this paper, we consider linear models:

$$f_j(\mathbf{a}) = \mathbf{w}_j^T \mathbf{a}, \quad j = 1, \ldots, k, \quad (1)$$

where $\mathbf{w}_j \in \mathbb{R}^n$ is the weight vector for the $j$-th task. The weight vectors for all $k$ tasks form the weight matrix $W = [\mathbf{w}_1, \ldots, \mathbf{w}_k] \in \mathbb{R}^{n \times k}$, which needs to be estimated from the data.

Assume that, given the value of $\mathbf{a}^j \in \mathbb{R}^n$, the corresponding target $y^j \in \mathbb{R}$ for the $j$-th task has a Gaussian distribution with mean $f_j(\mathbf{a}^j)$ and precision $\sigma^j > 0$ as

$$p(y^j|\mathbf{w}_j, \mathbf{a}^j, \sigma^j) = \sqrt{\frac{\sigma^j}{2\pi}} \exp\left(\frac{-\sigma^j(y^j - \mathbf{w}_j^T \mathbf{a}^j)^2}{2}\right). \quad (2)$$

Denote $\boldsymbol{\sigma} = [\sigma_1, \ldots, \sigma_k]^T \in \mathbb{R}^k$ and assume that the data $\{A, \mathbf{y}\}$ is drawn independently from the distribution in Eq. (2), then the likelihood function can be written as

$$p(\mathbf{y}|W, A, \boldsymbol{\sigma}) = \prod_{j=1}^k \prod_{i=1}^{m_j} p(y_i^j|\mathbf{w}_j, \mathbf{a}_i^j, \sigma^j). \quad (3)$$

To capture the task relatedness, a prior on $W$ is defined as follows. The $i$-th row of $W$, denoted as $\mathbf{w}^i \in \mathbb{R}^{1 \times k}$, corresponds to the $i$-th feature in all tasks. Assume that $\mathbf{w}^i$ is generated according to the exponential prior:

$$p(\mathbf{w}^i|\delta^i) \propto \exp(-\|\mathbf{w}^i\|\delta^i), \quad i = 1, 2, \ldots, n, \quad (4)$$

where $\delta^i > 0$ is the hyperparameter. Note that when $k = 1$, the prior in Eq. (4) reduces to the Laplace prior [8, 22].

Denote $\boldsymbol{\delta} = [\delta^1, \ldots, \delta^n]^T \in \mathbb{R}^n$, and assume that $\mathbf{w}^1, \ldots, \mathbf{w}^n$ are drawn independently from the prior in Eq. (4). Then the prior for $W$ can be expressed as

$$p(W|\boldsymbol{\delta}) = \prod_{i=1}^n p(\mathbf{w}^i|\delta^i). \quad (5)$$

It follows that the posterior distribution for $W$, which is proportional to the product of the prior and the likelihood function [4], is given by:

$$p(W|A, \mathbf{y}, \boldsymbol{\sigma}, \boldsymbol{\delta}) \propto p(\mathbf{y}|W, A, \boldsymbol{\sigma}) p(W|\boldsymbol{\delta}). \quad (6)$$

Taking the negative logarithm of Eq. (6) and combining with Eqs. (1-5), we obtain the maximum posterior estimation of $W$ by minimizing

$$\frac{1}{2} \sum_{j=1}^k \sigma^j \|\mathbf{y}_j - A_j \mathbf{w}_j\|^2 + \sum_{i=1}^n \delta^i \|\mathbf{w}^i\|. \quad (7)$$

For simplicity of discussion, we assume that $\sigma = \sigma^j, \forall j = 1, \ldots, k$ and $\delta = \delta^i, \forall i = 1, 2, \ldots, n$. We then obtain from Eq. (7) the following $\ell_{2,1}$-norm regularized least squares regression problem:

$$\min_W \frac{1}{2} \sum_{j=1}^k \|\mathbf{y}_j - A_j \mathbf{w}_j\|^2 + \rho \|W\|_{2,1}, \quad (8)$$

where $\rho = \delta/\sigma$ and $\|W\|_{2,1} = \sum_{i=1}^n \|\mathbf{w}^i\|$ is the $\ell_{2,1}$-norm of the matrix $W$.

**Discussion:** The problem in Eq. (8) can be generalized to the following $\ell_{2,1}$-norm regularization problem:

$$\min_{W \in \mathbb{R}^{n \times k}} \text{loss}(W) + \rho \|W\|_{2,1}, \quad (9)$$

where $\rho > 0$ is the regularization parameter, and $\text{loss}(W)$ is a smooth convex loss function such as the least square loss in Eq. (8) and the logistic loss.

When the number of tasks equals to one, the prior in Eq. (4) reduces to the Laplace prior distribution [8, 22]. It is easy to show that in this case the problem in Eq. (9) reduces to the $\ell_1$-norm regularized optimization problem. In particular, the problem in Eq. (8) reduces to the Lasso [23]. When there are multiple tasks, the weights corresponding to the $i$-th feature are grouped together via the $\ell_2$-norm of $\mathbf{w}^i$. Thus, the $\ell_{2,1}$-norm regularization tends to select features based on the strength of the input variables of the $k$ tasks jointly rather than on the strength of individual input variables as in the case of single task learning [1, 17, 18, 19].

## 3 Two Equivalent Smooth Reformulations

In this section, we propose to employ the Nesterov's method [14, 16] to solve the nonsmooth optimization problem in Eq. (9) by optimizing its equivalent smooth convex reformulations.

**Definition 1.** *The optimization problem*

$$\min_{\mathbf{x} \in G} g(\mathbf{x}) \quad (10)$$

*is called a constrained smooth convex optimization problem if $G$ is a closed convex set and $g(\mathbf{x})$ is a convex and smooth function defined in $G$.*

### 3.1 Constrained Smooth Convex Reformulations

The objective function in Eq. (9) is nonsmooth, since $\|W\|_{2,1}$ is non-differentiable. It is known that the subgradient method [14, 16] for solving this problem requires $O(\frac{1}{\varepsilon^2})$ iterations for achieving an accuracy of $\varepsilon$, which is significantly larger than $O(\frac{1}{\sqrt{\varepsilon}})$ required by the Nesterov's method for solving Eq. (10). We thus propose to solve the equivalent smooth convex reformulations of Eq. (9).

**The First Reformulation (aMTFL$_1$):** We introduce an additional variable $\mathbf{t} = [t_1, t_2, \ldots, t_n]^T$, where $t_i$ acts as the upper-bound of $\|\mathbf{w}^i\|$. With $\mathbf{t}$, we give the reformulation aMTFL$_1$ in the following theorem (proof given in Appendix A):

**Theorem 1.** *Let $\text{loss}(W)$ be a smooth convex loss function. Then the problem in Eq. (9) is equivalent to the following constrained smooth convex optimization problem:*

$$\min_{(\mathbf{t},W) \in D} \text{loss}(W) + \rho \sum_{i=1}^n t_i, \quad (11)$$

*where $\mathbf{t} = [t_1, t_2, \ldots, t_n]^T$ and*

$$D = \{(\mathbf{t}, W) | \|\mathbf{w}^i\| \le t_i, \forall i = 1, 2, \ldots, n\} \quad (12)$$

*is close and convex.*

**The Second Reformulation (aMTFL$_2$):** We can also derive an equivalent smooth reformulation by moving the nonsmooth $\ell_{2,1}$-norm term to the constraint. This results in the following equivalent reformulation aMTFL$_2$ (proof given in Appendix B):

**Theorem 2.** *Let $\text{loss}(W)$ be a smooth convex loss function. Then the problem in Eq. (9) is equivalent to the following $\ell_{2,1}$-ball constrained smooth convex optimization problem:*

$$\min_{W \in Z} \text{loss}(W), \quad (13)$$

*where*

$$Z = \{W \in \mathbb{R}^{n \times k} | \|W\|_{2,1} \le z\}, \quad (14)$$

$z \ge 0^1$ *is the radius of the $\ell_{2,1}$-ball, and there is a one-to-one correspondence between $\rho$ and $z$.*

**Remark:** For the reformulation aMTFL$_2$, it is difficult to give an analytical relationship between $\rho$ and $z$. However, such relationship is not critical, as in most cases the optimal values for $\rho$ and $z$ are both unknown, and they are usually tuned using cross-validation. It follows from Theorems 1 and 2 that we can solve Eq. (9) by the equivalent reformulations in Eqs. (11) and (13). For simplicity of presentation, we focus on solving the general problem (10), which is a constrained smooth convex optimization problem.

### 3.2 The Main Algorithm

We propose to employ the Nesterov's method [14, 16] for solving (10). Recall that the Nesterov's method has a much faster convergence rate than the traditional methods such as subgradient descent and gradient descent. Specifically, the Nesterov's method has a convergence rate of $O(\frac{1}{d^2})$, while gradient descent and subgradient descent have convergence rates of $O(\frac{1}{d})$ and $O(\frac{1}{\sqrt{d}})$, respectively, where $d$ denotes the number of iterations.

---
[1] In the following discussion, we skip the trivial case: $z = 0$.

The Nesterov's method is based on two sequences $\{\mathbf{x}_i\}$ and $\{\mathbf{s}_i\}$ in which $\{\mathbf{x}_i\}$ is the sequence of approximate solutions, and $\{\mathbf{s}_i\}$ is the sequence of search points. The search point $\mathbf{s}_i$ is the affine combination of $\mathbf{x}_{i-1}$ and $\mathbf{x}_i$ as

$$\mathbf{s}_i = \mathbf{x}_i + \alpha_i(\mathbf{x}_i - \mathbf{x}_{i-1}), \tag{15}$$

where $\alpha_i$ is the combination coefficient. The approximate solution $\mathbf{x}_{i+1}$ is computed as a "gradient" step of $\mathbf{s}_i$ as

$$\mathbf{x}_{i+1} = \pi_G(\mathbf{s}_i - \frac{1}{\gamma_i}g'(\mathbf{s}_i)), \tag{16}$$

where $\pi_G(\mathbf{v})$ is the Euclidean projection [5, Chapter 8.1, page 397] of $\mathbf{v}$ onto the convex set $G$:

$$\pi_G(\mathbf{v}) = \min_{\mathbf{x} \in G} \frac{1}{2}\|\mathbf{x} - \mathbf{v}\|^2, \tag{17}$$

$\frac{1}{\gamma_i}$ is the stepsize, and $\gamma_i$ is determined by the line search according to the Armijo-Goldstein rule so that $\gamma_i$ should be appropriate for $\mathbf{s}_i$ (details are given in Appendix F). A sequence of $\{\mathbf{x}_i\}$ and $\{\mathbf{s}_i\}$ is illustrated in Figure 1 in which $\mathbf{s}_i$ and $\mathbf{x}_{i+1}$ are computed recursively according to Eqs. (15) and (16) until the optimal solution $\mathbf{x}^*$ is obtained. The algorithm for solving Eq. (10) by the Nesterov's method is given in Algorithm 1. The definition of the function $g_{\gamma, \mathbf{s}_i}(\cdot)$ is given in Eq. (43) in Appendix F.

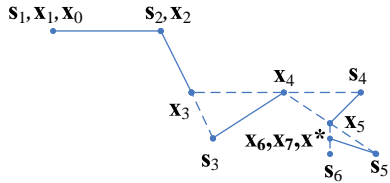

Figure 1: Illustration of Nesterov's method. We set $\mathbf{x}_1 = \mathbf{x}_0$ and $\alpha_2 = 0$, thus $\mathbf{s}_1 = \mathbf{x}_1$ and $\mathbf{s}_2 = \mathbf{x}_2$. The search point $\mathbf{s}_i$ is the affine combination of $\mathbf{x}_{i-1}$ and $\mathbf{x}_i$ (the dashed lines), and the next approximate solution is obtained by $\mathbf{x}_{i+1} = \pi_G(\mathbf{s}_i - \frac{1}{\gamma_i}g'(\mathbf{s}_i))$ (solid lines). We assume that $\mathbf{x}_6 = \mathbf{x}_7 = \mathbf{x}^*$, and $\mathbf{x}^*$ is an optimal solution.

**Theorem 3.** *[14, Theorem 11.3.1, page 176] When applying the Nesterov's method for solving the constrained smooth convex optimization problem in Eq. (10), one has*

$$g(\mathbf{x}_{d+1}) - \min_{\mathbf{x} \in G} g(\mathbf{x}) \leq \frac{2\hat{L}_g R_g^2}{(d+1)^2}, \tag{18}$$

*for any $d$, where $\hat{L}_g = \max(2L_g, L_0)$, $L_0$ is an initial guess of the Lipschitz continuous gradient $L_g$ of $g(.)$, and $R_g$ is the distance from the starting point to the optimal solution set.*

### 3.3 Time Complexity

We first analyze the time complexity of the problem in Eq. (11). When $\text{loss}(W)$ is either the least squares loss or the logistic loss, it costs $O(mn)$ floating point operations (flops) for evaluating the function value and gradient of the objective function of (11) at each iteration. Moreover, as shall be shown in Section 4.1, the Euclidean projections onto $D$ can be analytically computed in a time complexity of $O(nk)$. Therefore, through the equivalent reformulation (11), we can solve the $\ell_{2,1}$-norm regularized problem in Eq. (9) with a time complexity of $O(\frac{1}{\sqrt{\varepsilon}}(mn + nk))$, where $m$, $n$, $k$ and $\varepsilon$ denote the total number of training samples, the sample dimensionality, the number of tasks, and the desired accuracy, respectively. Following a similar analysis, the time complexity for solving Eq. (9) via the reformulation in Eq. (13) is also $O(\frac{1}{\sqrt{\varepsilon}}(mn + nk))$.

**Algorithm 1** Nesterov's Method for Constrained Smooth Convex Optimization
**Input:** $g(.), g'(.), G, L_0 > 0, \mathbf{x}_0$
**Output:** $\mathbf{x}$
1: Initialize $\mathbf{x}_1 = \mathbf{x}_0$, $t_{-1} = 0$, $t_0 = 1$, $\gamma_0 = L_0$
2: **for** $i = 1$ to ... **do**
3:    Set $\alpha_i = \frac{t_{i-2}-1}{t_{i-1}}$, $\mathbf{s}_i = \mathbf{x}_i + \alpha_i(\mathbf{x}_i - \mathbf{x}_{i-1})$
4:    **for** $j = 0$ to ... **do**
5:      Set $\gamma = 2^j \gamma_{i-1}$
6:      Compute $\mathbf{x}_{i+1} = \pi_G(\mathbf{s}_i - \frac{1}{\gamma}g'(\mathbf{s}_i))$
7:      **if** $g(\mathbf{x}_{i+1}) \leq g_{\gamma, \mathbf{s}_i}(\mathbf{x}_{i+1})$ **then**
8:         $\gamma_i = \gamma$, break
9:      **end if**
10:    **end for**
11:    Set $t_i = (1 + \sqrt{1 + 4t_{i-1}^2})/2$
12:    **if** convergence **then**
13:      $\mathbf{x} = \mathbf{x}_i$, terminate
14:    **end if**
15: **end for**

## 4 Efficient Euclidean Projections

A key building block of Algorithm 1 is the Euclidean projection onto the convex set $G$. In this section, we propose efficient algorithms for computing the Euclidean projection onto both $D$ in Eq. (12) and $Z$ in Eq. (14).

### 4.1 Euclidean Projection onto $D$

The Euclidean projection of a given point $(\mathbf{v}, U)$ onto the set $D$ is defined as:

$$\pi_D(\mathbf{v}, U) = \arg\min_{(\mathbf{t}, W) \in D} \frac{1}{2}\|W - U\|_F^2 + \frac{1}{2}\|\mathbf{t} - \mathbf{v}\|^2, \tag{19}$$

where $\mathbf{v}, \mathbf{t} \in \mathbb{R}^n$ and $U, W \in \mathbb{R}^{n \times k}$. Denote the $i$-th row of $U$ as $\mathbf{u}^i \in \mathbb{R}^{1 \times k}, i = 1, 2, \ldots, n$, and $\mathbf{v} = [v_1, v_2, \ldots, v_n]^T$. We show in the following theorem that the problem in Eq. (19) admits an analytical solution (proof given in Appendix C).

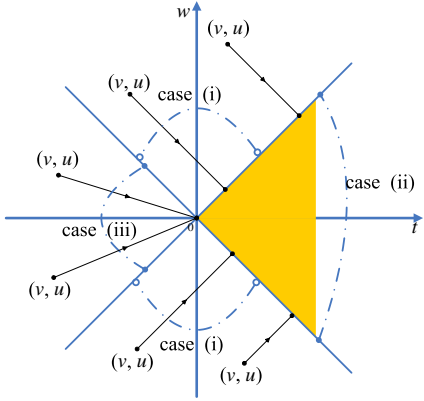

Figure 2: An illustration of the Euclidean projection onto the domain $D$ in the one-dimensional case. For convenience of illustration, we assume that $W$ is one-dimensional, and use the notations $w$ and $t$ for $W$ and $\mathbf{t}$, respectively. The domain $D$ corresponds to the golden region, i.e., $|w| \leq t$. The point to be projected $(v, u)$ can lie in the following three different regions: (i) $|u| > |v|$, (ii) $|u| \leq v$, and (iii) $|u| \leq -v$.

**Theorem 4.** *Denote the optimal solution to the problem in Eq. (19) by $\bar{W}$ and $\bar{\mathbf{t}}$, and the $i$-th row of $\bar{W}$ and $\bar{\mathbf{t}}$ by $\bar{\mathbf{w}}^i$ and $\bar{t}_i$, respectively. Then the problem has a unique solution given by:*

$$\bar{\mathbf{w}}^i = \begin{cases} \dfrac{\|\mathbf{u}^i\| + v_i}{2\|\mathbf{u}^i\|}\mathbf{u}^i & \|\mathbf{u}^i\| > |v_i|, \\ \mathbf{u}^i & \|\mathbf{u}^i\| \leq v_i, \\ \mathbf{0} & \|\mathbf{u}^i\| \leq -v_i, \end{cases} \quad (20)$$

$$\bar{t}_i = \begin{cases} \dfrac{\|\mathbf{u}^i\| + v_i}{2} & \|\mathbf{u}^i\| > |v_i|, \\ v_i & \|\mathbf{u}^i\| \leq v_i, \\ 0 & \|\mathbf{u}^i\| \leq -v_i. \end{cases} \quad (21)$$

The Euclidean projection in Eq. (19) can be interpreted from a geometrical perspective as shown in Figure 2. We denote the point to be projected as $(v, u)$. When $(v, u)$ lies in the region of case (i), its Euclidean projection onto the domain $D$ is $\left(\frac{|u|+v}{2|u|}u, \frac{|u|+v}{2}\right)$, as can be observed from the four illustrating points in Figure 2. When $(v, u)$ lies in the region of case (ii), its Euclidean projection onto the domain $D$ is still $(v, u)$, as it already resides in the domain $D$. When $(v, u)$ lies in the region of case (iii), the nearest point in $D$ to $(v, u)$ (in terms of the Euclidean distance) is the origin $(0, 0)$.

We analyze the time complexity of this projection as follows. It costs $O(nk)$ flops to compute $\|\mathbf{u}^i\|$, for $i = 1, 2, \ldots, n$, and $O(nk)$ flops to obtain $\bar{W}$ and $\bar{\mathbf{t}}$ from Eqs. (20) and (21). Therefore, the Euclidean projection onto $D$ can be computed in $O(nk)$ time.

### 4.2 Euclidean Projection onto $Z$

The Euclidean projection of $U \in \mathbb{R}^{n \times k}$ onto the domain $Z$ is defined as:

$$\begin{aligned}\pi_Z(U) = \quad & \arg\min_W \quad \frac{1}{2}\|W - U\|_F^2, \\ & \text{s.t.} \quad \|W\|_{2,1} \leq z. \end{aligned} \quad (22)$$

We use the standard Lagrangian method to solve the problem in Eq. (22). Introducing the Lagrangian variable $\lambda$ for the inequality constraint $\|W\|_{2,1} \leq z$, we can write the Lagrangian of Eq. (22) as:

$$L(W, \lambda) = \frac{1}{2}\|W - U\|_F^2 + \lambda(\|W\|_{2,1} - z). \quad (23)$$

Let $\bar{W}$ be the primal optimal point and $\bar{\lambda}$ be the dual optimal point and denote the $i$-th row of $\bar{W}$ by $\bar{\mathbf{w}}^i \in \mathbb{R}^k$. It is clear that the primal and dual optimal points satisfy the following conditions:

$$\|\bar{W}\|_{2,1} \leq z, \quad (24)$$
$$\bar{\lambda} \geq 0. \quad (25)$$

In Eq. (22), both the objective and the constraints are convex. When $W$ is a zero matrix, we have $\|W\|_{2,1} - z < 0$ since $z > 0$. Therefore, the Slater's condition [5] for strong duality holds and the complementary slackness conditions [5] follows:

$$\bar{\lambda}(\|\bar{W}\|_{2,1} - z) = 0. \quad (26)$$

We show in the following theorem (proof given in Appendix D) that when the dual optimal point $\bar{\lambda}$ is known, the primal optimal point $\bar{W}$ can be computed analytically.

**Theorem 5.** *When the dual optimal point $\bar{\lambda}$ is known, the primal optimal point $\bar{W}$ is given by*

$$\bar{\mathbf{w}}_i = \begin{cases} \left(1 - \dfrac{\bar{\lambda}}{\|\mathbf{u}^i\|}\right)\mathbf{u}^i, & \bar{\lambda} > 0, \|\mathbf{u}^i\| > \bar{\lambda} \\ \mathbf{0}, & \bar{\lambda} > 0, \|\mathbf{u}^i\| \leq \bar{\lambda} \\ \mathbf{u}^i, & \bar{\lambda} = 0, \end{cases} \quad (27)$$

*for $i = 1, \ldots, n$, where $\mathbf{u}^i \in \mathbb{R}^{1 \times k}$ is the $i$-th row of $U$.*

Next, we show how to compute the dual optimal $\bar{\lambda}$ in the following theorem (proof given in Appendix E).

**Theorem 6.** *For the Euclidean projection in Eq. (22), if $\|U\|_{2,1} \leq z$, the dual optimal is given by*

$$\bar{\lambda} = 0, \quad (28)$$

*and if $\|U\|_{2,1} > z$, the dual optimal $\bar{\lambda} > 0$ is the unique root of the following function:*

$$\omega(\lambda) = \sum_{i=1}^{n} \max(\|\mathbf{u}^i\| - \lambda, 0) - z. \quad (29)$$

Theorems 5 and 6 suggest the following procedure for solving Eq. (22). First, we compute the dual optimal $\bar{\lambda}$, which is either zero or the unique root of Eq. (29); and then we analytically obtain $\bar{W}$ from Eq. (27). We employ the improved bisection [12] for computing the root of $\omega(\cdot)$ in Eq. (29).

We analyze the time complexity of this projection as follows. It costs $O(nk)$ flops to compute $\|\mathbf{u}^i\|$, for $i = 1, 2, \ldots, n$, $O(n)$ flops to compute $\bar{\lambda}$ as the root of Eq. (29) by the improved bisection [12], and $O(nk)$ flops for obtaining $\bar{W}$ by Eq. (27). Therefore, the overall time complexity for solving Eq. (22) is $O(nk)$, which scales linearly with the size of $U \in \mathbb{R}^{n \times k}$ to be projected.

## 5 Experiments

In this section, we demonstrate the efficiency of the proposed algorithms using the following two data sets: School[2] and Letter [17, 18]. The School data set consists of scores of 15,362 students from 139 secondary schools in London during the years 1985, 1986, and 1987, with each sample containing 28 attributes. The School data set has 139 tasks of predicting student performance in each school. The Letter data set was collected by Rob Kassel at the MIT Spoken Language Systems Group. This data set consists of 8 default tasks of two-class classification problems for the handwritten letters: c/e, g/y, m/n, a/g, i/j, a/o, f/t, h/n. The writings are collected from over 180 different writers, with the letters being represented by $8 \times 16$ binary pixel images. The Letter data set has a total number of 45,679 samples.

**Convergence Analysis** We solve the $\ell_{2,1}$-norm regularized least squares regression problem (8) by the proposed algorithms in Section 3, and report the results on both data sets (including all samples) in Figure 3. The horizontal axis denotes the number of iterations and the vertical axis denotes the objective function value. For the reformulation aMTFL$_1$, we report results for $\rho = 0.1$ and $\rho = 0.001$ on the School data set; and $\rho = 1$ and $\rho = 0.1$ on the Letter data set. For the reformulation aMTFL$_2$, we set $z = \|\bar{W}\|_{2,1}$, where $\bar{W}$ is the solution obtained by aMTFL$_1$. We can observe form the figures: i) the objective function values for both reformulations decrease rapidly at the first few iterations; and ii) the function values for both reformulations become stable after about 30 iterations with the specific $\rho$ and $z$ on these two data sets. This confirms the fast convergence rate of the prosposed algorithms due to the use of the Nesterov's method.

**Comparison with Competing Algorithms** To demonstrate the efficiency of the proposed algorithms in Section 3, we compare them with the gradient descent (GD) that solves the reformulation in Eq. (13). We also compare the proposed algorithms with MTL-FEAT[3] [1] for solving

---

[2] www.mlwin.com/intro/datasets.html
[3] www.cs.ucl.ac.uk/staff/A.Argyriou/code

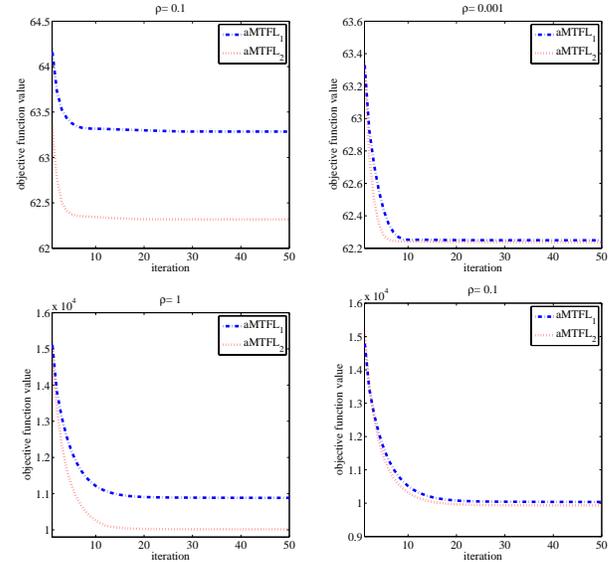

Figure 3: Illustration of the proposed algorithms for solving the $\ell_{2,1}$-norm regularized least squares problem: plots in the first and second row correspond to the results on the School and Letter data sets, respectively; aMTFL$_1$ and aMTFL$_2$ correspond to reformulations (11) and (13), respectively.

the squared $\ell_{2,1}$-norm regularized least squares regression given by [1, Eq. (5) of Section 2]:

$$\min_W \sum_{j=1}^k \|\mathbf{y}_j - A_j \mathbf{w}_j\|^2 + \gamma \|W\|_{2,1}^2, \quad (30)$$

which is slightly different from the optimization problem in Eq. (8) discussed in this paper. Before the comparison, we note that it is very difficult, if not possible, to carry out a fair comparison among the different methods, due to the issue of implementation, the choice of algorithm parameters, and the different stopping criterion, and thus the results should be interpreted with caution. To conduct a relatively fair comparison, for a given $\gamma$, we first run the MTL-FEAT to obtain the solution $\bar{W}$; then apply the algorithms proposed in Section 3 for solving (11) with $\rho = \gamma \|\bar{W}\|_{2,1}/2$ and (13) with $z = \|\bar{W}\|_{2,1}$. For MTL-FEAT, we use the following settings: "epsilon_init=0", "iterations=20", and "method=3". For our proposed algorithms and GD, we terminate the program after they achieve an objective function value smaller or equal to that of MTL-FEAT in term of Eq. (30). On the School data set, we test the efficiency under different values of the regularization parameter $\gamma$, and report the results in Table 1; on the Letter data set, we test the efficiency with a varying number of training samples (denoted by $m$) under a given $\gamma$ (set to 0.001), and report the results in Table 2. We can observe from these two tables that the proposed algorithms are much more efficient than both MTL-FEAT and GD.

| $\gamma$ | 0.1 | 0.01 | 0.001 | 0.0001 |
|---|---|---|---|---|
| MTL-FEAT | 17.90 | 18.01 | 18.02 | 18.26 |
| aMTFL$_1$ | 0.04 | 0.13 | 0.76 | 2.89 |
| aMTFL$_2$ | 0.06 | 0.22 | 0.60 | 1.32 |
| GD | 0.07 | 0.62 | 4.60 | 6.63 |

Table 1: Comparison of the computational time (in seconds) among MTL-FEAT, the proposed two reformulations, and GD for the $\ell_{2,1}$-norm regularized least squares regression on the School data set.

| $m$ | 916 | 1832 | 2748 | 3664 | 4580 |
|---|---|---|---|---|---|
| MTL-FEAT | 1.11 | 5.84 | 16.84 | 36.79 | 67.99 |
| aMTFL$_1$ | 0.03 | 0.03 | 0.12 | 0.18 | 0.32 |
| aMTFL$_2$ | 0.07 | 0.18 | 0.36 | 0.44 | 0.82 |
| GD | 0.41 | 0.91 | 1.97 | 1.98 | 2.41 |

Table 2: Comparison of the computational time (in seconds) among MTL-FEAT, the proposed two reformulations, and GD for the $\ell_{2,1}$-norm regularized least squares regression on the Letter data set.

The superior performance of both aMTFL$_1$ and aMTFL$_2$ over MTL-FEAT can be explained as follows. At each iteration, MTL-FEAT forms $k$ kernel matrices (of sizes $m_j \times m_j, j = 1, 2, \ldots, k$) and computes the Singular Value Decomposition (SVD) of these kernel matrices, which costs $O(\sum_{j=1}^k m_j^3 + \sum_{j=1}^k m_j^2 n)$ flops; while aMTFL$_1$ and aMTFL$_2$ do not involve SVD, and only cost $O(mn + nk)$ flops at each iteration. Thus, aMTFL$_1$ and aMTFL$_2$ are much more efficient than MTL-FEAT, especially when the total sample size $m = \sum_{j=1}^k m_j$ is large.

**aMTFL$_1$ Versus aMTFL$_2$** We compare the two reformulations aMTFL$_1$ and aMTFL$_2$ for solving the $\ell_{2,1}$-norm regularized least squares regression. For a given regularization parameter $\rho$, we first run aMTFL$_1$ to obtain the solution $\bar{W}$, where the program is terminated when the relative gap of the adjacent approximate solutions is less than $10^{-4}$; and then run aMTFL$_2$ with $z = \|\bar{W}\|_{2,1}$, where the program is terminated when it achieves an objective function value smaller than or equal to that of aMTFL$_1$ in terms of Eq. (11). We conduct experiments on the School data with different values of the regularization parameter $\rho$, and report the results in Table 3. We can observe from the table that, when $\rho$ is relatively large, aMTFL$_1$ is comparable to (or more efficient than) aMTFL$_2$; and when $\rho$ is relatively small, aMTFL$_2$ is more efficient than aMTFL$_2$. Similar phenomenon can be observed from Table 1.

Next, we analyze the underlying difference between these two reformulations based on the parameters involved in (18) of Theorem 3: i) the Lipschitz gradient of $\text{loss}(W) + \rho \sum_{i=1}^n t_i$ is identical to that of $\text{loss}(W)$, thus both reformulations have the same Lipschitz gradient $L_g$; ii) due to the additional variable $\mathbf{t}$, aMTFL$_1$ usually has a larger $R_g$ (the distance from the starting point to the optimal solution set) than aMTFL$_2$; and iii) from the discussion in Section 4, the computation of the Euclidean projection for aMTFL$_1$ is less expensive than that for aMTFL$_2$.

| $\rho$ | 1.5 | 1 | 0.8 | 0.1 | 0.01 | 0.001 |
|---|---|---|---|---|---|---|
| aMTFL$_1$ | 0.13 | 0.14 | 0.16 | 0.76 | 2.48 | 6.29 |
| aMTFL$_2$ | 0.15 | 0.16 | 0.18 | 0.46 | 1.26 | 1.90 |

Table 3: Comparison of the computational time (in seconds) between aMTFL$_1$ and aMTFL$_2$ for the $\ell_{2,1}$-norm regularized least squares regression on the School data set.

When $\rho$ is relatively large, the solution is sparse. Thus, the value of $R_g$ for aMTFL$_1$ is comparable to that for aMTFL$_2$. This, together with the efficient computation of the Euclidean projection for aMTFL$_1$, explains why aMTFL$_1$ is competitive with aMTFL$_2$ when $\rho$ is relatively large. When $\rho$ is relatively small, the solution is less sparse. In this case, the value of $R_g$ for aMTFL$_1$ is much larger than that for aMTFL$_2$. This explains why aMTFL$_1$ is less efficient than aMTFL$_2$ in this case.

**Computing a Sequence of Solutions** In most applications, the optimal values for $\rho$ and $z$ are both unknown. One common approach for estimating these valies is to compute the solutions corresponding to a sequence of values of the parameter $\rho_1 > \rho_2 > \ldots > \rho_s$ or $z_1 < z_2 < \ldots < z_s$, from which the optimal one is chosen by evaluating certain criteria. This can be done by simply applying the proposed algorithms to solving the $s$ independent problems (called "cold-start"). However, a more efficient approach is through the so-called "warm-start", which uses the solution of the previous problem as the warm-start of the latter. Indeed, both aMTL$_1$ and aMTL$_2$ can benefit from the warm-start technique, since the solution corresponding to $\rho_i$ (or $z_i$) lies in the feasible domain of $\rho_{i+1}$ (or $z_{i+1}$). We conduct experiments on the School data set and report results in Figure 4. The horizontal axis denotes a sequence of values of the parameter $\rho_1 > \rho_2 > \ldots > \rho_s$ (left plot) or $z_1 < z_2 < \ldots < z_s$ (right plot), and the vertical axis denotes the number of iterations for computing the solution for a given $\rho_i$ (or $z_i$) with the cold-start and warm-start. For the cold-start, all problems are solved independently; for the warm-start, the solution corresponding to $\rho_i$ (or $z_i$) is employed as the initialization for the solution corresponding to $\rho_{i+1}$ (or $z_{i+1}$). From the results in this figure, we can observe that the warm-start can significantly outperform the code-start.

## 6 Conclusion

In this paper, we study the problem of joint feature selection across a group of related tasks. We consider the multi-task feature learning formulation based on the $\ell_{2,1}$-norm regularization, which encourages multiple predictors to share similar sparsity patterns. The resulting optimization problem is, however, challenging to solve due to the non-smoothness of the $\ell_{2,1}$-norm. To accelerate the computation, we propose to firstly reformulate it as two equivalent smooth convex optimization problems, and then solve the reformulations by the Nesterov's method. In solving the reformulations, a key building block is the Euclidean projection. We show that the Euclidean projection for the first reformulation can be analytically computed, while the projection for the second one can be computed in linear time. Empirical evaluations on several data sets demonstarte the efficiency of the proposed algorithms.

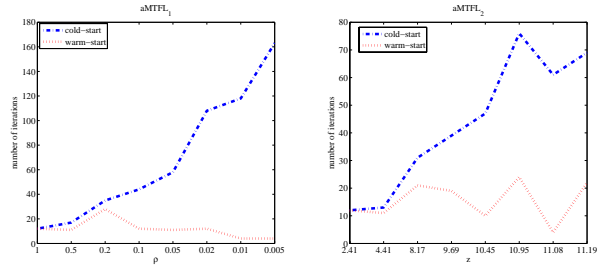

Figure 4: Solving a sequence of solutions for aMTFL$_1$ and aMTFL$_2$ by cold-start and warm-start: for the cold-start, all problems are solved independently; for the warm-start, the solution corresponding to $\rho_i$ (or $z_i$) is employed as the initialization for the solution corresponding to $\rho_{i+1}$ (or $z_{i+1}$).

We plan to improve the practical performance of the proposed algorithms using the adaptive line search proposed in [11], and apply the proposed algorithms for solving the $\ell_{2,1}$-norm regularized problems including the group Lasso [27], the logistic group-lasso [13], and the group-lasso for generalized linear models [21]. We also plan to compare the proposed algorithms with the coordinate gradient descent method [25], which has a locally linear convergence rate under certain conditions [25, Theorem 4]. Finally, we plan to apply the proposed algorithms to real-world applications including association analysis of quantitative trait network [9], image classification [20], brain-computer interfacing [24], and biomarker identification [10].

**Acknowledgements**

We would like to thank the anonymous reviewers for helpful comments. This work was supported by NSF IIS-0612069, IIS-0812551, CCF-0811790, NIH R01-HG002516, and NGA HM1582-08-1-0016.## References

[1] A. Argyriou, T. Evgeniou, and Massimiliano Pontil. Convex multi-task feature learning. *Machine Learning*, 73(3):243–272, 2008.

[2] S. Ben-david and R. Schuller. Exploiting task relatedness for multiple task learning. In *Computational Learning Theory*, 2003.

[3] J. Bi, T. Xiong, S. Yu, M. Dundar, and B. Rao. An improved multi-task learning approach with applications in medical diagnosis. In *European Conference on Machine Learning*, 2008.

[4] C. M. Bishop. *Pattern Recognition and Machine Learning*. Springer, 2006.

[5] S. Boyd and L. Vandenberghe. *Convex Optimization*. Cambridge University Press, 2004.

[6] J. Dattorro. *Convex Optimization & Euclidean Distance Geometry*. Meboo Publishing, 2005.

[7] T. Evgeniou and M. Pontil. Regularized multi-task learning. In *ACM SIGKDD International Conference on Knowledge discovery and data mining*, pages 109–117, 2004.

[8] J. Goodman. Exponential priors for maximum entropy models. In *The Annual Meeting of the Association for Computational Linguistics*, pages 305–312, 2004.

[9] S. Kim, K.-A. Sohn, and E. P. Xing. A multivariate regression approach to association analysis of quantitative trait network. Technical Report CMU-ML-08-113, Carnegie Mellon University, 2008.

[10] S. Kim and E. Xing. Feature selection via block-regularized regression. In *Uncertainty in Artificial Intelligence*, 2008.

[11] J. Liu, J. Chen, and J. Ye. Large-scale sparse logistic regression. In *ACM SIGKDD International Conference On Knowledge Discovery and Data Mining*, 2009.

[12] J. Liu and J. Ye. Efficient euclidean projections in linear time. In *International Conference on Machine Learning*, 2009.

[13] L. Meier, S. Geer, and P. Bühlmann. The group lasso for logistic regression. *Journal of the Royal Statistical Society: Series B*, 70:53–71, 2008.

[14] A. Nemirovski. *Efficient methods in convex programming*. Lecture Notes, 1994.

[15] A. Nemirovski and D. Yudin. *Problem Complexity and Method Efficiency in Optimization*. John Wiley and Sons, 1983.

[16] Y. Nesterov. *Introductory Lectures on Convex Optimization: A Basic Course*. Kluwer Academic Publishers, 2003.

[17] G. Obozinski, B. Taskar, and M. I. Jordan. Multi-task feature selection. Technical report, Statistics Department, UC Berkeley, 2006.

[18] G. Obozinski, B. Taskar, and M. I. Jordan. Joint covariate selection for grouped classification. Technical report, Statistics Department, UC Berkeley, 2007.

[19] G. Obozinski, M. J. Wainwright, and M. I. Jordan. High-dimensional union support recovery in multivariate regression. In *Neural Information Processing Systems*, 2008.

[20] A. Quattoni, X. Carreras, M. Collins, and T. Darrell. A projected subgradient method for scalable multi-task learning. Technical report, MIT-CSAIL-TR-2008-045, 2008.

[21] V. Roth and B. Fischer. The group-lasso for generalized linear models: uniqueness of solutions and efficient algorithms. In *International conference on Machine learning*, pages 848–855, 2008.

# Appendix

### Appendix A. Proof of Theorem 1

We begin with a lemma, which states that $D$ defined in (12) is a closed convex set.

**Lemma 1.** *The set $D$ defined in (12) is closed and convex.*

**Proof**: Denote

$$D_i = \{(t_i, \mathbf{w}^i) | \|\mathbf{w}^i\| \le t_i\}, i = 1, 2, \ldots, n, \quad (31)$$

which is known as the ice-cream cone [5, 6]. Since the ice-cream cone is closed and convex, $D_i$ is a closed convex cone. As the Cartesian product of closed convex cones is a closed convex cone [6, Chapter 2, page 148], we conclude that

$$D = D_1 \times D_2 \times \ldots \times D_n \quad (32)$$

is a closed convex cone, and a closed convex set as well. □

We are now ready to prove Theorem 1. The objective function of (11) is smooth and convex, as both $\text{loss}(W)$ and $\rho \sum_{i=1}^n t_i$ are smooth and convex. Moreover, from Lemma 1, $D$ is a closed convex set. Therefore, (11) is a constrained smooth convex optimization problem. The problem in (9) is equivalent to (11), since at the optimal point of (11), we have $t_i = \|\mathbf{w}^i\|, \forall i = 1, 2, \ldots, n$.

### Appendix B. Proof of Theorem 2

We first prove that (13) is a constrained smooth convex optimization problem. It is easy to verify that $\|W\|_{2,1}$ is a norm. Since any norm is a closed convex function [16, Example 3.1.1, Chapter 3.1.1, page 107], we conclude that $\|W\|_{2,1}$ is a closed convex function. As the sublevel set [16, Theorem 3.1.3, page 106] of a closed convex function is either empty or a closed convex set, we conclude that

$$Z = \{W | \|W\|_{2,1} \le z\} \quad (33)$$

is closed and convex (note that, $Z$ is not empty, since $z > 0$ and the zero matrix belongs to $Z$). We conclude that (13) is a constrained smooth convex optimization problem. The equivalence relationship between (9) and (13) follows from the Lagrangian duality.

### Appendix C. Proof of Theorem 4

We begin with a lemma, which gives the well-known variational inequality (v.i.) condition for the optimal solution of the constrained smooth convex optimization.

**Lemma 2.** *[16, Chapter 2.2, Theorem 2.2.5, page 79] For the constrained smooth convex optimization problem (10), the point $\mathbf{x}^* \in G$ is a solution to (10) if and only if*

$$\langle \mathbf{x} - \mathbf{x}^*, f'(\mathbf{x}^*) \rangle \ge 0, \forall \mathbf{x} \in G. \quad (34)$$

Next, we prove Theorem 4. The objective function of (19) is strictly convex, and the domain $D$ is closed and convex (see Theorem 1), and thus the Euclidean projection problem in (19) has a unique solution. It is easy to observe that, the problem Eq. (19) can be decoupled into the following subproblems:

$$\min_{(t_i, \mathbf{w}^i) \in D_i} \frac{1}{2}\|\mathbf{w}^i - \mathbf{u}^i\|^2 + \frac{1}{2}\|t_i - v_i\|^2 \quad (35)$$

for $i = 1, 2, \ldots, n$, where $D_i$ is defined in (31). It is clear that (35) has a unique solution.

From Lemma 2, we only need to verify that the solution given in (20) and (21) belongs to the feasible domain $D_i$ and meanwhile satisfies the v.i. condition of the problem (35). It is easy to verify that, the solution given in (20) and (21) satisfies $\|\bar{\mathbf{w}}_i\| \le \bar{t}_i$, so that it belong to $D_i$.

Next, we verify that the solution given in (20) and (21) satisfies the v.i. condition for (35), i.e., $\forall \|\mathbf{w}^i\| \le t_i$,

$$\langle \mathbf{w}^i - \bar{\mathbf{w}}^i, \bar{\mathbf{w}}^i - \mathbf{u}^i \rangle + (t_i - \bar{t}_i)(\bar{t}_i - v_i) \ge 0. \quad (36)$$

We consider the following three cases for $(v_i, \mathbf{u}^i)$: (i) $\|\mathbf{u}^i\| > |v_i|$; (ii) $\|\mathbf{u}^i\| \le v_i$; and (iii) $\|\mathbf{u}^i\| \le -v_i$.

For the first case, we denote $\boldsymbol{\alpha} = \frac{v_i - \|\mathbf{u}^i\|}{2\|\mathbf{u}^i\|}\mathbf{u}^i$ and $\beta =$

$\frac{\|\mathbf{u}^i\| - v_i}{2}$. It follows that $\|\boldsymbol{\alpha}\| = \beta > 0$. We have

$$\begin{aligned}
&\langle \mathbf{w}^i - \bar{\mathbf{w}}_i, \bar{\mathbf{w}}_i - \mathbf{u}^i \rangle + (t_i - \bar{t}_i)(\bar{t}_i - v_i) \\
&= \langle \mathbf{w}^i, \boldsymbol{\alpha} \rangle + t_i \beta \\
&\geq -\|\mathbf{w}^i\| \cdot \|\boldsymbol{\alpha}\| + t_i \beta \\
&\geq -t_i \cdot \beta + t_i \beta = 0,
\end{aligned}$$

where the first equality follows from (20) and (21), the first inequality follows from the property that $|\langle \mathbf{w}^i, \boldsymbol{\alpha} \rangle| \leq \|\mathbf{w}_i\| \cdot \|\boldsymbol{\alpha}\|$, the second inequality follows from $\|\mathbf{w}^i\| \leq t_i$ and $\|\boldsymbol{\alpha}\| = \beta$. Thus, the v.i. condition (36) holds.

For the second case, the v.i. condition (36) automatically holds, since $\bar{\mathbf{w}}^i = \mathbf{u}^i$ and $\bar{t}_i = v_i$.

For the third case, we have

$$\begin{aligned}
&\langle \mathbf{w}^i - \bar{\mathbf{w}}_i, \bar{\mathbf{w}}_i - \mathbf{u}^i \rangle + (t_i - \bar{t}_i)(\bar{t}_i - v_i) \\
&= \langle \mathbf{w}^i, -\mathbf{u}^i \rangle + t_i(-v_i) \\
&\geq -\|\mathbf{w}^i\| \cdot \|\mathbf{u}^i\| - t_i v_i \\
&\geq -t_i \cdot (-v_i) - t_i v_i = 0,
\end{aligned}$$

where the first inequality utilizes $|\langle \mathbf{w}^i, \mathbf{u}^i \rangle| \leq \|\mathbf{w}^i\| \cdot \|\mathbf{u}^i\|$, and the second inequality follows from $v_i < 0, \|\mathbf{u}^i\| \leq -v_i$ and $\|\mathbf{w}^i\| \leq t_i$. Thus, the v.i. condition (36) holds.

**Appendix D. Proof of Theorem 5**

It is clear that $\bar{W}$ is the optimal solution to

$$\bar{W} = \arg\min_W L(W, \bar{\lambda}), \tag{37}$$

which can be decoupled into

$$\bar{\mathbf{w}}^i = \arg\min_{\mathbf{w}^i} h(\mathbf{w}^i) \equiv \frac{1}{2}\|\mathbf{w}^i - \mathbf{u}^i\|^2 + \bar{\lambda}\|\mathbf{w}^i\|. \tag{38}$$

Since $h(\mathbf{w}^i)$ is strictly convex, we conclude that $\bar{\mathbf{w}}^i$ is its unique minimizer. It is known that, a point is the optimal solution to an unconstrained nonsmooth convex optimization problem if and only if the zero point belongs to its subdifferential [16, Theorem 3.1.15, Chapter 3.1, page 119]. That is, $\bar{\mathbf{w}}^i$ is the optimal solution to (38) if and only if

$$\mathbf{0} \in \partial h(\bar{\mathbf{w}}^i), \tag{39}$$

where

$$\partial h(\mathbf{w}^i) = \mathbf{w}^i - \mathbf{u}^i + \bar{\lambda}\partial\|\mathbf{w}^i\|, \tag{40}$$

$$\partial\|\mathbf{w}^i\| = \begin{cases} \left\{\frac{\mathbf{w}^i}{\|\mathbf{w}^i\|}\right\} & \mathbf{w}^i \neq \mathbf{0} \\ \{\mathbf{y} \in \mathbb{R}^{1\times k} | \|\mathbf{y}\| \leq 1\} & \mathbf{w}^i = \mathbf{0}. \end{cases} \tag{41}$$

The computation of $\partial\|\mathbf{w}^i\|$ can be found in [16, Chapter 3.1, page 122]. Next, we verify that the solution given in (27) satisfies the subdifferential condition in (39).

When $\bar{\lambda} > 0$ and $\|\mathbf{u}^i\| > \bar{\lambda}$, we have $\bar{\mathbf{w}}^i = (1 - \frac{\bar{\lambda}}{\|\mathbf{u}^i\|})\mathbf{u}^i$ according to (27). We have $\partial h(\bar{\mathbf{w}}^i) = \{\mathbf{0}\}$, since

$$\partial\|\bar{\mathbf{w}}^i\| = \left\{\frac{\bar{\mathbf{w}}^i}{\|\bar{\mathbf{w}}^i\|}\right\} = \left\{\frac{\mathbf{u}^i}{\|\mathbf{u}^i\|}\right\}.$$

When $\bar{\lambda} > 0$ and $\|\mathbf{u}^i\| \leq \bar{\lambda}$, we have $\bar{\mathbf{w}}^i = \mathbf{0}$ according to (27). From (41), we have $\mathbf{u}^i/\bar{\lambda} \in \partial\|\bar{\mathbf{w}}^i\|$, which leads to

$$\bar{\mathbf{w}}^i - \mathbf{u}^i + \bar{\lambda}\mathbf{u}^i/\bar{\lambda} = 0. \tag{42}$$

Thus, (39) holds.

When $\bar{\lambda} = 0$, we have $\bar{\mathbf{w}}^i = \mathbf{u}^i$ according to (27). It follows from (40) that (39) holds. This completes the proof.

**Appendix E. Proof of Theorem 6**

We first show that $\bar{\lambda} = 0$, if $\|U\|_{2,1} \leq z$. Otherwise, from (25), we have $\bar{\lambda} > 0$, which leads to $\|\bar{W}\|_{2,1} = z$, by using the slackness complementary condition (26). From (27), we have $\|\bar{\mathbf{w}}^i\| \leq \|\mathbf{u}^i\|$, where the equality holds if and only if $\mathbf{u}^i = \mathbf{0}$ (note that, we have assumed $\bar{\lambda} > 0$). Therefore, we have $\|\bar{W}\|_{2,1} \leq \|U\|_{2,1}$, where the equality holds if and only if $\mathbf{u}^i = \mathbf{0}, \forall i = 1, 2, \ldots, n$. From $\|U\|_{2,1} \leq z$, $\|\bar{W}\|_{2,1} = z$ and $\|\bar{W}\|_{2,1} \leq \|U\|_{2,1}$, we have $\|\bar{W}\|_{2,1} = \|U\|_{2,1} = z > 0$ and $\mathbf{u}^i = \mathbf{0}, \forall i = 1, 2, \ldots, n$, leading to a contradiction. Therefore, we have $\bar{\lambda} = 0$.

Next, we show that $\bar{\lambda} > 0$, if $\|U\|_{2,1} > z$. Otherwise, from (25), we have $\bar{\lambda} = 0$, which leads to $\bar{W} = U$ from (27). We have $\|\bar{W}\|_{2,1} = \|U\|_{2,1} > z$, which contradicts with (24). Therefore, $\bar{\lambda} > 0$. If follows from the slackness complementary condition (26) that

$$\|\bar{W}\|_{2,1} - z = 0,$$

which leads to $\omega(\bar{\lambda}) = 0$ by using (27).

Finally, we show that the function $\omega(\lambda)$ has a unique root. It is easy to verify that, $\max(\|\mathbf{u}^i\| - \lambda, 0)$ is continuous and monotonically decreasing in $(-\infty, +\infty)$, and strictly decreasing in $(-\infty, \|\mathbf{u}^i\|]$. Therefore, $\omega(\lambda)$ is continuous and monotonically decreasing in $(-\infty, +\infty)$, and strictly decreasing in $(-\infty, \max_i \|\mathbf{u}^i\|]$. We have $\omega(0) = \|U\|_{2,1} - z > 0$ and $\omega(\max_i \|\mathbf{u}^i\|) = -z < 0$. According to the Intermediate Value Theorem, $\omega(\lambda)$ has a unique root, which lies in $(0, \max_i \|\mathbf{u}^i\|)$.

**Appendix F. Definition of an Appropriate $\gamma$**

The choice of $\gamma$ is key for the convergence analysis of the Nesterov's method. For a given $\gamma > 0$, we define

$$g_{\gamma,\mathbf{x}}(\mathbf{y}) = g(\mathbf{x}) + \langle g'(\mathbf{x}), \mathbf{y} - \mathbf{x} \rangle + \frac{\gamma}{2}\|\mathbf{y} - \mathbf{x}\|^2, \tag{43}$$

which is the tangent line of $g(.)$ at $\mathbf{x}$, regularized by the square distance between $\mathbf{y}$ and $\mathbf{x}$. It is easy to verify that the quadratic function $g_{\gamma,\mathbf{x}}(\mathbf{y})$ is strongly convex, and minimizing $g_{\gamma,\mathbf{x}}(\mathbf{y})$ in the domain $G$ is equivalent to the problem of Euclidean projections onto $G$, that is,

$$\pi_G(\mathbf{x} - \frac{1}{\gamma}g'(\mathbf{x})) = \arg\min_{\mathbf{y} \in G} g_{\gamma,\mathbf{x}}(\mathbf{y}). \tag{44}$$

We say that $\gamma$ is "appropriate" for $\mathbf{x}$, if the following holds:

$$g(\pi_G(\mathbf{x} - \frac{1}{\gamma}g'(\mathbf{x}))) \leq g_{\gamma,\mathbf{x}}(\pi_G(\mathbf{x} - \frac{1}{\gamma}g'(\mathbf{x}))). \tag{45}$$